%% file: arxiv.tex
\documentclass[11pt,final]{article}
\usepackage{coling2020}
\usepackage{times}
\usepackage{url}
\usepackage{latexsym}

\newcommand{\ignore}[1]{}
\newcommand{\nop}[1]{}

\newcommand{\eg}{\textit{e.g.,~}}
\newcommand{\ie}{\textit{i.e.,~}} 

\usepackage{array}

\usepackage[utf8x]{inputenc}

\usepackage{subcaption,enumitem}

\usepackage{multirow}

\usepackage{tikz}
\usepackage{pgfplots} 
\usepgfplotslibrary{polar}
\usepackage{xcolor,colortbl}

\usepgfplotslibrary{fillbetween}

\usepackage{xcolor,colortbl}   

\colorlet{expFourheatmap}{red!45!white}%\definecolor{expFourheatmap}{rgb}{255,0,0}%{255,0,0}
\colorlet{expFiveheatmap}{red!45!white}%\definecolor{expFiveheatmap}{rgb}{255,0,0}

\colorlet{f1expFour}{blue!45!white}

\colorlet{f1expFive}{blue!45!white}

\colorlet{correct}{blue!50}%orange} 
\colorlet{incorect}{red!50}%orange} 

\colorlet{TE}{black} 
\colorlet{TEc}{orange!75!black} 
\colorlet{TEw}{orange!50!white} 

\colorlet{TETEcTEw}{cyan}

\colorlet{BertTE}{blue!50!white}%!75!black} 
\colorlet{BertTEc}{blue!50!white}%!75!black} 
\colorlet{BertTEw}{blue!50!white}%!75!black} 
\colorlet{BertTETEcTEw}{blue!75!black} 

\colorlet{BertTE3}{black!50!white} 
\colorlet{BertTETEcTEw3}{black!80!white} 
\colorlet{BertTE4}{red!50!white} 
\colorlet{BertTETEcTEw4}{red!70!black} 
 
\colorlet{BertTE}{black!80!white}%!75!black} 
\definecolor{BertTEc}{RGB}{55,126,184}%{255,0,0}{red!50!white}%!75!black} 
\definecolor{BertTEw}{RGB}{77,175,74}%{255,0,0}{violet!50!white}%!75!black} 
\colorlet{BertTETEcTEw}{black!50!white}%!75!black} 

\colorlet{ElmoTE}{BertTE}%red!50!white}%!75!black} 
\colorlet{ElmoTEc}{BertTEc}%{red!50!white}%!75!black} 
\colorlet{ElmoTEw}{BertTEw}%{red!50!white}%!75!black} 
\colorlet{ElmoTETEcTEw}{BertTETEcTEw}%{red!75!black} 

\colorlet{GloveTE}{BertTE}%green!75!black!50!white} 
\colorlet{GloveTEc}{BertTEc}%{green!75!black!50!white} 
\colorlet{GloveTEw}{BertTEw}%{green!75!black!50!white} 
\colorlet{GloveTETEcTEw}{BertTETEcTEw}%{green!30!black} 

\tikzstyle{linestyleTE}= [solid, thick]%only marks]%solid]
\tikzstyle{linestyleTEc}= [solid, thick]%only marks]%dotted]
\tikzstyle{linestyleTEw}= [solid, thick]%only marks]%dashed]
\tikzstyle{linestyleTETEcTEw}= [solid, thick]%only marks]%dashdotted]

\tikzstyle{linestyleBertTE}=[linestyleTE] 
\tikzstyle{linestyleBertTEc}= [linestyleTEc] 
\tikzstyle{linestyleBertTEw}= [linestyleTEw] 
\tikzstyle{linestyleBertTETEcTEw}=[linestyleTETEcTEw] 

\tikzstyle{linestyleElmoTE}= [linestyleTE]
\tikzstyle{linestyleElmoTEc}= [linestyleTEc]
\tikzstyle{linestyleElmoTEw}= [linestyleTEw]
\tikzstyle{linestyleElmoTETEcTEw}= [linestyleTETEcTEw]

\tikzstyle{linestyleGloveTE}= [linestyleTE]
\tikzstyle{linestyleGloveTEc}= [linestyleTEc]
\tikzstyle{linestyleGloveTEw}= [linestyleTEw]
\tikzstyle{linestyleGloveTETEcTEw}= [linestyleTETEcTEw]

\tikzstyle{lineplotMarkTE}= [mark=*, mark size=0.75, mark options={solid, thin}] 
\tikzstyle{lineplotMarkTEc}= [mark=*, mark size=0.75, mark options={solid,thin}]  
\tikzstyle{lineplotMarkTEw}= [mark=*, mark size=0.75, mark options={solid,thin}] 
\tikzstyle{lineplotMarkTETEcTEw}= [mark=*, mark size=0.75, mark options={solid,thin}]

\colorlet{radarTEc}{cyan}%orange!75!black} 
\colorlet{radarTEw}{red}%violet}%red} 
\tikzstyle{radarplotMarkTEc}= [mark=*, mark options={solid,thick}]  
\tikzstyle{radarplotMarkTEw}= [mark=square*, mark options={solid,thick}]  
\tikzstyle{radarplotFillTEc}= [fill=radarTEc, fill opacity=0.2]  
\tikzstyle{radarplotFillTEw}= [fill=radarTEw, fill opacity=0.2]  

\colorlet{radarTE}{black}%orange!75!black} 
\colorlet{radarTEmixed}{purple}%violet}%red} 
\tikzstyle{radarplotMarkTEc}= [mark=x, mark options={solid,thick}]  
\tikzstyle{radarplotMarkTEw}= [mark=diamond*, mark options={solid,thick}]  
\tikzstyle{radarplotFillTE}= [fill=radarTE, fill opacity=0.2]  
\tikzstyle{radarplotFillTEmixed}= [fill=radarTEmixed, fill opacity=0.2]

\colorlet{crosslangAllLangModel}{orange}%!50!white}%orange!75!black} 
\colorlet{crosslangPerLangModel}{orange}%red} 
\tikzstyle{crosslangAllLangModelbar}= [fill=crosslangAllLangModel, fill opacity=0.9]  
\tikzstyle{crosslangPerLangModelbar}= [fill=crosslangPerLangModel, fill opacity=0.3]

\colorlet{ratioHMneg}{blue}%orange} 
\colorlet{ratioHMpos}{orange}%blue} 

\tikzstyle{ethLinestyle}= [dash pattern=on 3pt off 2pt]
\tikzstyle{xmrLinestyle}= [dash pattern=on \pgflinewidth off 1pt]
\usepackage{tkz-kiviat} 
\usepackage{csvsimple}

\usetikzlibrary{arrows}
\colorlet{crosslangAllLangModel}{orange}%!50!white}%orange!75!black} 
\colorlet{crosslangPerLangModel}{orange}%red} 
\tikzstyle{crosslangAllLangModelbar}= [fill=crosslangAllLangModel, fill opacity=0.9]  
\tikzstyle{crosslangPerLangModelbar}= [fill=crosslangPerLangModel, fill opacity=0.3]

\newcommand{\new}[1]{#1}

\colingfinalcopy % Uncomment this line for the final submission

\title{Towards Trustworthy Deception Detection: Benchmarking Model Robustness across Domains, Modalities, and Languages}

\author{Maria Glenski, Ellyn Ayton, Robin Cosbey, Dustin Arendt, and   Svitlana Volkova \\ 
  Pacific Northwest National Laboratory \\
  Richland, WA, USA \\
  {\tt first.last@pnnl.gov} \\}

\date{}

\begin{document}
\maketitle

\begin{abstract}
Evaluating model robustness 
is critical when developing trustworthy models not only to gain deeper understanding of model behavior, strengths, and weaknesses, but also to develop future models that are generalizable and robust across expected environments a model may encounter in deployment. 
In this paper we present a framework for measuring model robustness for an important but difficult text classification task -- deceptive news detection. 
We evaluate model robustness to out-of-domain data, modality-specific features, and languages other than English.

Our investigation focuses on three type of models: LSTM models trained on multiple datasets (Cross-Domain), several fusion LSTM models trained with images and text and evaluated with three state-of-the-art embeddings, BERT ELMo, and GloVe (Cross-Modality), and character-level CNN models trained on multiple languages (Cross-Language).
Our analyses reveal a significant drop in performance when testing neural
models on out-of-domain data and non-English languages that may be mitigated using diverse training data. We find
that with additional image content as input, ELMo embeddings yield significantly fewer errors compared to BERT or GLoVe. Most importantly, this work not only carefully analyzes deception model robustness but also provides a framework of these analyses that can be applied to new models or extended datasets in the future. 

\end{abstract}

\blfootnote{This work is licensed under a Creative Commons Attribution 4.0 International License.} 
\blfootnote{License details:  \url{http://creativecommons.org/licenses/by/4.0/}.} 

%%%%%%%%%%%%%%%%%%%%%%%%%%%%%%%%%%%%
% Main Text
%%%%%%%%%%%%%%%%%%%%%%%%%%%%%%%%%%%

\section{Introduction}

Detection of deceptive content online is an extremely important but challenging task and there have been significant efforts to apply machine learning and deep learning  
to solve this problem~\cite{rubin2016fake,mitra2017parsimonious,wang2017liar,karadzhov2017we,volkova2017separating,shu2017fake,rashkin2017truth}. 
 A plethora of models exist that rely on a range of data \new{mined} from various social platforms -- Facebook, Twitter, Reddit -- and modalities -- text, images, or both. These existing models rely on different text features (\eg linguistic structure, lexical, psycholinguistic features, biased and subjective language), image features, and user interaction features (\eg engagement, network structure, temporal patterns). However, current literature lacks details for how these models transfer, \eg to out-of-domain data, across platforms, or across languages.  

Moreover, there is a gap in understanding the underlying behavior of the decision-making processes behind model output. It is not clear why models for deception detection (particularly neural or deep learning based models) are making certain predictions (\eg \textit{why} one item is predicted as deceptive versus not). Evaluation of model performance with one metric such as accuracy or F1 score is not enough. For such complex prediction tasks like deception detection for which humans often disagree~\cite{karduni2018can,karduni2019vulnerable,ott2011finding,harris2012detecting}, we need more rigorous evaluation of neural model behavior and a cohesive system for comparing model results across bodies of research.

Evaluations need to explicitly measure the extent to which model performance is affected when new data is supplied, \eg data with a different topic distribution or  
how well-performing models on English data will perform on non-English inputs. 
There is also a critical need for evaluations that highlight when a model is correct, which examples have the ability to explain why, and, most importantly, conclude why a user should trust a given model \new{in an interpretable manner}. Arguments in favor of the above requirements to model performance are well-aligned with recent work on machine learning interpretability, trust, fairness, accountability, and reliability~\cite{lipton2016mythos,doshi2017towards,hohman2018visual}.  
Another key argument in favor of rigorous evaluation is the lack of benchmark datasets and the need for reproducibility. Often, researchers do not make models or datasets publicly available and the details to reconstruct their models and experimental setup are not sufficient, which prevents other researchers from performing rigorous  
comparisons. To reproduce our key findings and experiments, we will make our framework available through interactive Jupyter notebooks available at publication.
 
The focus of the current work and our 
main contribution is an {\it extensive evaluation of neural model robustness across frequently encountered factors in real-world applications of digital deception models} such as new domains, languages, 
modalities, and upgrades to new state-of-the-art text embeddings.

\section{Related Work}
 
Deep learning systems have been adopted in many areas from medicine to autonomous driving \cite{ahmad2018interpretable,claybrook2018autonomous} and as these algorithms are incorporated, 
the need for explainable and transparent models becomes more urgent. One approach researchers use to overcome the inherent ambiguity of these black-box methods is to develop additional models to learn and explain the decisions of existing models, analyze when these models fail, and introduce a human-in-the-loop component to improve performance \cite{ribeiro2016should,murdoch2019interpretable,poursabzi2018manipulating}. 
Another way to tackle this challenge is to design models with a goal of interpretability in place when  development starts~\cite{ridgeway1998interpretable,rudin2018please,gilpin2018explaining,lahav2018interpretable,hooker2019benchmark}. 
Model performance beyond accuracy, precision, and recall measures are essential in order to build trustworthy and reliable models tasked with making decisions capable of significantly affecting the end users \cite{hohman2018visual,dodge2019explaining,hohman2019gamut}.

In this study, we do not focus on developing new visualization techniques to explain deception detection models using intrinsic or post-hoc explanations~\cite{yang2019xfake,shu2019defend,reis2019explainable,wallace2019allennlp} or measure their interpretability~\cite{mohseni2019open}. Instead, we thoroughly evaluate model performance using error analysis under several real-world conditions e.g., across domains and input types to understand reliability and the underlying behavior of decision making processes that will in turn increase model explainability and interpretability.

{\bf Deception Detection}  
Detecting suspicious or deceptive news  
online is a well explored area of study. Current research efforts focus on broadly classifying between suspicious and trustworthy content \cite{volkova2017separating,shu2020hierarchical} to more specific distinctions such as between propaganda, hoax, satire, and trustworthy news \cite{rashkin2017truth}, to examining the behavior of malicious users and bots \cite{glenski2018humans,kumar2017army,kumar2018rev2}, or analyzing misinformation and rumor spread patterns over time \cite{kwon2017rumor,vosoughi2018spread}.  
Methods for classification vary from random forests to deep neural networks and the addition of enriched features such as images, temporal and structural attributes, and linguistic features have been shown to boost model performance over relying on textual characteristics alone \cite{wang2017liar,qazvinian2011rumor,kwon2013prominent}. 

With the high impact of digital deception on offline, real-world events, effective detection models are a critical concern. However, interpretable evaluations of model performance beyond traditional  precision, recall, or F score measures are essential to building trustworthy and reliable systems when deep learning or machine learning models are tasked with making decisions capable of significantly affecting end users \cite{dodge2019explaining,volkova2019explaining,hohman2019gamut,hohman2018visual}.

Many studies have relied on Recurrent Neural Networks (RNNs) or variations,  
particularly Long Short Term Memory (LSTM) layers, for deception detection tasks~\cite{ma2016detecting,chen2018unsupervised,rath2017retweet,zubiaga2018discourse,zhang2019reply}. 
Others have used Convolutional Neural Networks (CNN)~\cite{ajao2018fake},  
or variations of LSTM architectures such as including attention mechanisms~\cite{guo2018rumor,li2019rumor} which are typically dependent on specific tasks or parameter tuning of state-of-the-art deception detection models. For the purposes of consistency across our experiments, we rely on standard LSTM models that underpin much of the recent work rather than tuning each architecture by task and data or comparing the wider range of recent state-of-the-art architectures. This enables us to make more accurate comparisons of the factors related to robustness that we vary in our experiments. 
Developing novel models  
or comparing all state-of-the-art architectures is beyond the scope of this paper.

{\bf Evaluating Model Robustness}
Measuring robustness of a model goes beyond fortifying against intentional manipulations. Corrupted data inputs are possible in real-world scenarios without harboring the intent to fool a system and models should behave as expected in such cases, \eg simple image transformations, missing or misspelled text. Several key studies have endeavored to develop methods to evaluate model robustness \cite{zheng2016improving,hein2017formal,liu2018towards,dodge2019work}.
In this study we focus on evaluating model robustness to out-of-domain data, multilingual data, and multimodal inputs combined with conceptually different text embedding techniques to gain deeper insights into how much ``learning and understanding`` neural network models actually have. Different studies have separately examined the relationships between extracted features across
modalities, across languages, and across domains~\cite{zhou2020safe,capuozzo2020decop,zotova2020multilingual}.
To the best of our knowledge, this extensive evaluation of deception detection model robustness across multiple tasks has not been presented before in one unified work.

\section{Methodology} 
 
In this section we discuss the state-of-the-art neural models for
deception detection used in this paper and  
describe our approach for evaluating model robustness across domains, modalities, and languages, a summary of which is presented in Table~\ref{tab:overview}.

\begin{table*}[b]
	\small
	\centering
	\input{robustness_only_figs/robustness_overview_table} 
	\caption{Overview of the deception detection models ($M_1$, $M_2$, $M_3$, $M_4$, and $M_5$) used for each task.} 
	\label{tab:overview}
\end{table*}
 
\subsection{Neural Models for Deception Detection} \label{Neural Models}

After an extensive analysis of published neural deception classification models, we adopt similar architectures that consist of a commonly used RNN  
architecture,  
LSTM layers, and rely on combinations of text, linguistic cue (LC), and image vectors for our  
 experiments. 
 Lexical vectors and linguistic cues include those often used for classification, notably encoding for biased and otherwise subjective language~\cite{rashkin2017truth,shu2019beyond}. We employ LIWC~\cite{pennebaker2001linguistic} and several lexical dictionaries (assertive verbs, hedges, factives, implicatives, etc.~\cite{recasens2013linguistic}) to construct frequency vectors for model input.  
Image vectors are representations extracted from the last layer of the state-of-the-art ResNet architecture~\cite{he2016deep}. 
We focus our analysis on both models with single modalities \eg text and multimodal models \eg text and image, as well as classification tasks over a varied number of classes. We create 80/20 splits in each dataset for train and test sets. 
 
We implement the binary classification model over trustworthy and deceptive samples using text and lexical input features to evaluate the efficacy and robustness of testing on out-of-domain data. 
We additionally implement the binary (trustworthy versus deceptive), the 3-way (trustworthy, propaganda, disinformation), and the 4-way (clickbait, satire, hoax, conspiracy) classification models with the enhancement of image input features for our evaluation of multimodal models. Our final model to assess the robustness of these deep learning approaches is the 3-way classification of clickbait, conspiracy, and propaganda news sources across multiple languages. 
We employ similar neural network 
architectures for all models, which consists of a pre-trained text embedding layer followed by an LSTM layer and a single dense layer. The output from this sub-network is concatenated to the image vectors which have been transformed by two dense layers. In the architectures of our multi-modal models, our joint text and image representations are concatenated to the lexical cues vector as input to the final two fully connected layers in the model. All layers use dropout.\footnote{For each model, we perform a random hyper-parameter tuning search independently consisting of ten different configurations of model hyper-parameters. 
The batch size, the number of training epochs, drop out rate, and the recurrent layer dimensionality as well as the optimizer and learning rate are among the tunable parameters. 
Every model had a unique set of final hyper-parameters, however, the most common configurations consisted of the Adam optimizer with a learning rate of 0.0001, a drop out rate between 0.2 and 0.25, and 10 epochs of training.}  

In contrast to the Cross-Domain and Cross-Modality models, the multilingual Cross-Language models do not incorporate linguistic cues or other lexical features because of the inconsistency across multiple languages -- some of the linguistic features and lexicons are not available for all languages (\eg bias and subjective language dictionaries).
The multilingual model ($M_5$) architecture is
constructed by the pre-trained character-level text embedding layer followed by 
two convolution layers, then a max pooling layer, and finally two
dense, fully connected layers.

\subsection{Datasets}

\begin{table}[t!]
	\small
	\centering
	\begin{tabular}{l@{\hspace{1pt}}r@{\hspace{5pt}}r@{\hspace{5pt}}r@{\hspace{5pt}}r@{\hspace{5pt}}}
		& \multicolumn{2}{c}{\bf Train } &  \multicolumn{2}{c}{\bf Test }\\
		Data&  \it Trustworthy &  \it Deceptive & \it  Trustworthy &  \it Deceptive \\  
		\hline 
		Twitter & 14.9k & 28.0k & 3.7k & 7.2k \\  
		Reddit & 21.6k & 21.5k & 5.4k & 5.5k \\ 
		Twitter + Reddit & 36.6k & 49.9k & -- & -- \\
		
		\hline
	\end{tabular}
	\vspace{-0.5\baselineskip}
	\caption{Distributions across classes within train and test data used in the cross-domain analyses ($M_1$).}
	\label{tab:cross_domains_breakdown}
\end{table}

We use  
a previously released, public list of news sources\footnote{We use the following publicly available news source annotations for \textit{deceptive news sources}: \url{https://www.cs.jhu.edu/~svitlana/data/SuspiciousNewsAccountList.tsv} and \textit{trustworthy news sources}: \url{https://www.cs.jhu.edu/~svitlana/data/VerifiedNewsAccountList.tsv}} annotated along a spectrum of deceptive %media
-- clickbait, hoax, satire, conspiracy, propaganda, or disinformation --  and verified news sources who typically spread factual content (which we denote as ``trustworthy'' in our experiments). These annotated news sources focused on activity in 2016, therefore, we collected the following
datasets for the same 12 month period of activity. %we collected data to cover this same time period.
Note, there are limitations when annotations are done on the sources level, however, similar to related work~\cite{vosoughi2018spread,lazer2018science,glenski2018propagation} we advocate focusing on the news
sources rather than individual stories because we view the definitive element of deception to be the intent and the tactics of the news 
source. 

 \vspace{0.2\baselineskip} 
In our {\it \textbf{Cross-Domain}} analyses, we consider two domains -- Twitter and Reddit. The Twitter component of the $M_1$ dataset is comprised of English retweets from official twitter accounts for news media described above, containing only text-based posts. 
The Reddit component comprises all top-level comments in response to posts on Reddit in 2016 where the post contained a link to a web domain associated with one of the news accounts of interest. We create a binary classification dataset, analogous to the Twitter dataset used for $M_2$, that collapses news source annotations to either trustworthy or deceptive and down-samples the Reddit comments to have approximately the same volume of content ($N=54k$). This allows the joint {\it Twitter + Reddit} dataset to be balanced between the two domains for cross-domain analyses, as shown in Table~\ref{tab:cross_domains_breakdown} where we present the class distributions for the $M_1$ dataset.

 \vspace{0.2\baselineskip} 
Our multimodal Twitter data, for the  {\it \textbf{Cross-Modality}} analyses, is comprised of 
the same collection of English retweets from annotated news media's official accounts used to build the Twitter component of the $M_1$ dataset,  
filtered to those tweets that include images. Each retweet comprises a unique image and body of text. We create datasets for three classification tasks: (1) propaganda, disinformation, or trustworthy ($M_3$; $N=54.4k$), (2) clickbait, conspiracy, hoax, or satire ($M_4$; $N=2.5k$), and (3) a binary classification dataset ($M_2$; $N=54.4k$) of trustworthy versus deceptive tweets in the dataset used for $M_3$ by collapsing the sub-classes of propaganda and disinformation into a single deceptive class. The breakdown of these three tasks is presented in Table~\ref{tab:overview}.

 \vspace{0.2\baselineskip} 
In contrast to the previous, English-only datasets, our multilingual Twitter dataset for the  {\it \textbf{Cross-Language}} analyses comprises 7,316 tweets across five languages (English, French, German, Russian, and Spanish) from a similar collection of tweets as described above for English Twitter without the restriction to only English-text. Because of an over-representation of English-tweets in the original collection, we sample 1,500 examples for each language evenly distributed across the three classes, \ie 500 clickbait samples, 500 conspiracy samples, and 500 propaganda samples. For the least represented language, Russian, we use all available tweets: 478 clickbait samples, 338 conspiracy samples, and 500 propaganda samples. For all languages, we allocate 80\% to train and 20\% to test.

\subsection{Evaluating Neural Model Robustness} 

We investigate the robustness of digital deception models when evaluated across three dimensions: domains, modalities and languages, seeking to answer three key research questions.
Table \ref{tab:overview} outlines the model architectures, inputs, datasets, and prediction tasks for all robustness analyses.

 \vspace{0.5\baselineskip} 
  In our {\it \textbf{Cross-Domain}} analyses, we evaluate model robustness across two popular social platforms: Twitter and Reddit. We compare model performance on a binary classification task ($M_1$ in Table~\ref{tab:overview}) using the binary English Twitter and Reddit datasets summarized in Table~\ref{tab:cross_domains_breakdown}. We train three LSTM models using textual and lexical input features -- (1) trained on Twitter content only, (2) trained on Reddit content only, and (3) trained on both the Twitter and Reddit content. All three models rely on pre-trained BERT word embeddings that are fine-tuned during training and use the same neural architecture and hyper-parameters described in section \ref{Neural Models}. 
  To account for inherent platform inconsistencies between Twitter and Reddit, we limit the input text length to 100 words and use a common vocabulary between the three model setups.
  We evaluate each model on two held out test sets composed of held out examples from Twitter and Reddit allowing us to test performance within the same domain (\eg train and test on Twitter data) and on out-of-domain data (\eg train on Twitter and test on Reddit).

\vspace{0.5\baselineskip} 
 In our {\it \textbf{Cross-Modality}} analyses, we evaluate how multiple modalities can influence the predictions of a model. We train three LSTM classifiers to predict falsified news at different levels of granularity. The first model ($M_2$) is trained to distinguish between trustworthy and deceptive news using text, lexical cues, and extracted image features. The second model ($M_3$) uses more fine-grained labels of digital deception to further separate deceptive news into propaganda and disinformation. Finally, ($M_4$) is a 4-way classification task to predict tweets from four types of deceptive news types: clickbait, conspiracy, hoax, and satire. Each classifier is trained three separate times using a different text embedding strategy that allows for fine-tuning during training: BERT, ELMo, or GloVe.
 We use the Tensorflow Object Detection API~\cite{huang2017speed} to extract the primary COCO\footnote{\url{http://cocodataset.org/#home}} (Common Objects in COntext) objects from the images of the tweets. In this analysis, we also seek to answer, in particular, the question: {\it Which errors do text embeddings or detected objects contribute to in the multimodal setup?}

\vspace{0.5\baselineskip} 
 In our {\it \textbf{Cross-Language}} analyses, we evaluate changes in model performance when a convolutional neural network ($M_5$) developed to distinguish fine-grained differences in digital deception (\textit{clickbait} versus \textit{conspiracy} versus \textit{propaganda}) is trained and evaluated across multiple languages. We examine the impacts on performance when the model is trained in the context of multiple languages (\ie a single model trained using an aggregated dataset of English, Russian, German, Spanish, and French samples) and evaluated on a single language as well as when the model is trained in the context of a single language and tested on the same language using 5-fold cross-validation.

\section{Experimental Results}
In this section, we highlight the results of our experiments regarding model robustness on out-of-domain data (Cross-Domain analyses), across multimodal inputs using various text embeddings (Cross-Modality analyses), and multilingual inputs (Cross-Language analyses).

\begin{figure}[t!] 
	\centering 
	\small
	%\begin{tabular}{p{0.31\textwidth}}
	``RT fluoridated water now linked to diabetes \& lowered IQ still drinking it''
	%\end{tabular} 
	
	\includegraphics[width=0.2\textwidth, clip,keepaspectratio]{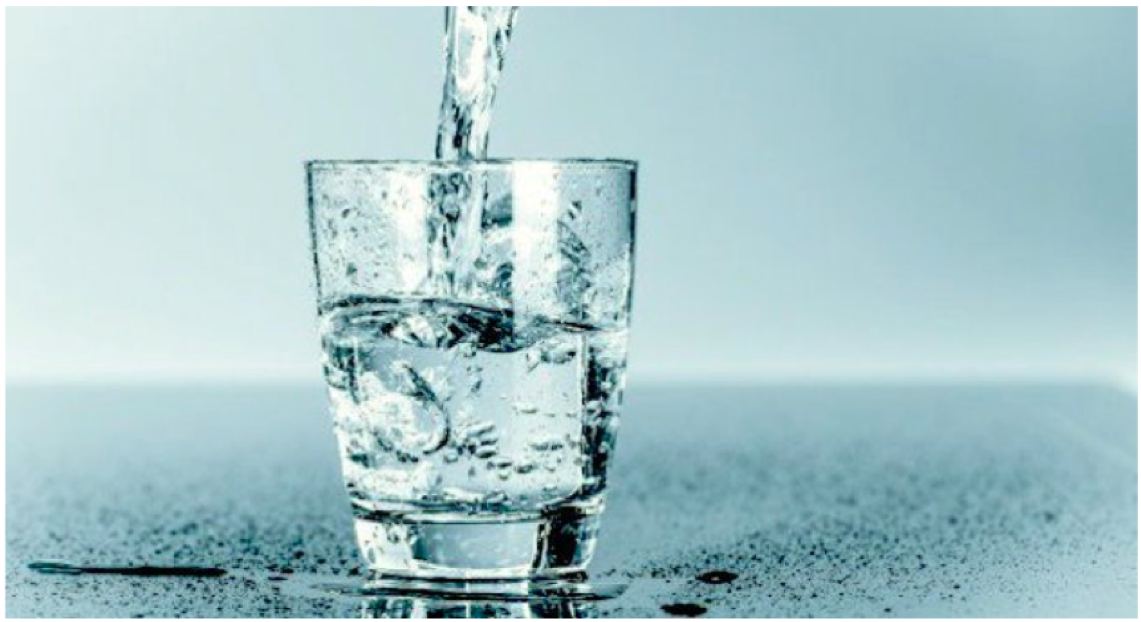}
	  
	\caption{Incorrect classification with high confidence example: a tweet annotated \textit{Conspiracy}  
		classified as \textit{Clickbait} with 92.6\% confidence by the $M_4$ (ELMo embeddings) model.}
	\label{example1}  
\end{figure}

Before 
we do so,
we manually validate the models qualitatively by examining incorrectly classified tweets with high confidences. 
For example, the 4-way classification
 model, $M_4$, that relies on pre-trained ELMo embeddings shows the best performance, but makes errors that resemble human mistakes. 
 The tweet shown in Figure~\ref{example1} is an example of a high-confidence misclassification that could intuitively be similarly made by a human annotator. From the text or image alone the true classification is not immediately clear or intuitive from a human standpoint. The text implies conspiracy, but it is not obvious from the image  -- a stock image of a glass of water being poured -- to which class it should belong. This analysis of high-confidence incorrect classifications provides a better understanding of how high confidence misclassifications (especially high-confidence which can serve as a proxy for where end-users \textit{should} -- or would feel that they should -- trust the model) occur and under what conditions.

\subsection{Cross-Domain Performance}

\begin{figure}[t!]
	\input{figs/exp1confusionmatrix} 
	\vspace{-\baselineskip}
	\caption{Confusion matrices for cross-domain analyses using models that classify content as Trustworthy or Deceptive. 
		Cells are shaded by the rate of (mis)classification; % shown in parentheses. 
		Correct predictions are shaded in blue,  misclassifications in red.}
	\label{fig:exp1confusionmatrices}
\end{figure}

We present the confusion matrices for the cross-domain experiments in Figure~\ref{fig:exp1confusionmatrices} and, as expected, we observe the best individual performance when models are trained and tested on same-domain (same-platform) data. Additionally, we see that performance suffers on models tested on out-of-domain data (models trained on Twitter but test on Reddit and vice-versa). However, we find that both of the cross-domain models ($M_1$) trained on either the Twitter or Reddit data alone over-predict the \textit{trustworthy} class when tested on out-of-domain test data. This is irrespective of the distribution of class instances (shown in Table~\ref{tab:cross_domains_breakdown}) -- whether it is imbalanced (as it is in Twitter) or not (as in Reddit).  
As illustrated in the final row of confusion matrices, we see that the best overall performance is achieved by the model with \textit{domain diverse} training data -- trained on both the Twitter and Reddit training datasets combined. This model {\it performs comparably on both the Twitter and Reddit held out test sets and as well or better than the individually trained models when tested on in-domain samples.} 

Our findings confirm that data from new domains (\ie social platforms) has a high impact on model performance even in the cases where classes are defined in the same way (as is the case with our Twitter and Reddit datasets that rely on annotations of the same pool of news sources as trustworthy or deceptive). 
On the Twitter test data, the domain-diverse model is even able to more accurately identify trustworthy content (60\% true positives compared to 46\% when trained on Twitter data alone) and reduce the number of false positive predictions of Deceptive content (40\% of trustworthy tweets classified as deceptive compared to 54\%). 
This indicates that even if used only for a single platform, engaging diverse training data examples across multiple platforms can result in a more robust, more trustworthy model.

\subsection{Cross-Modality Performance}

\begin{figure}[t]
    \centering
    
    \includegraphics[height=8.5\baselineskip]{ 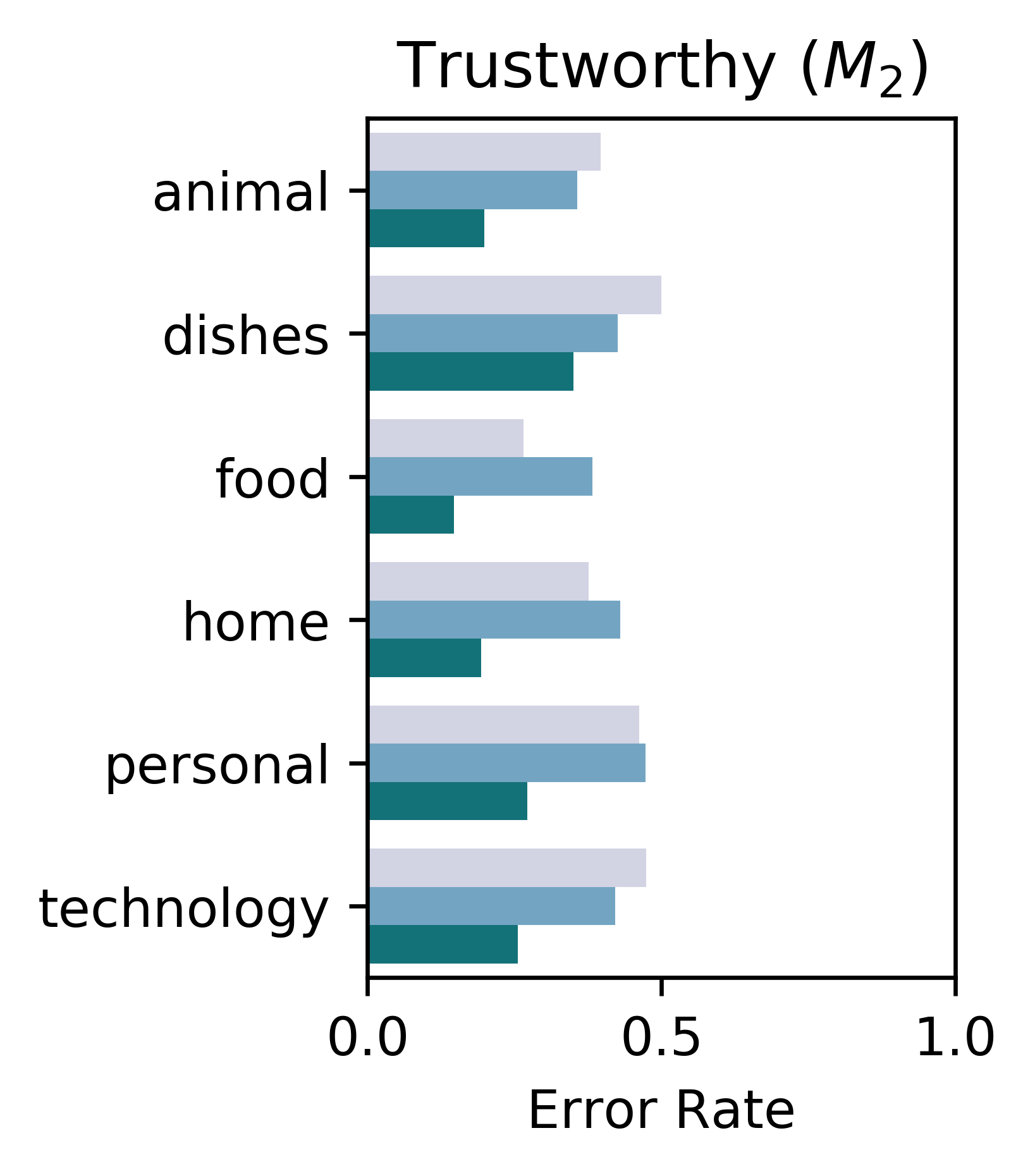}
    \hspace{-0.8\baselineskip}
    \includegraphics[height=8.5\baselineskip]{ 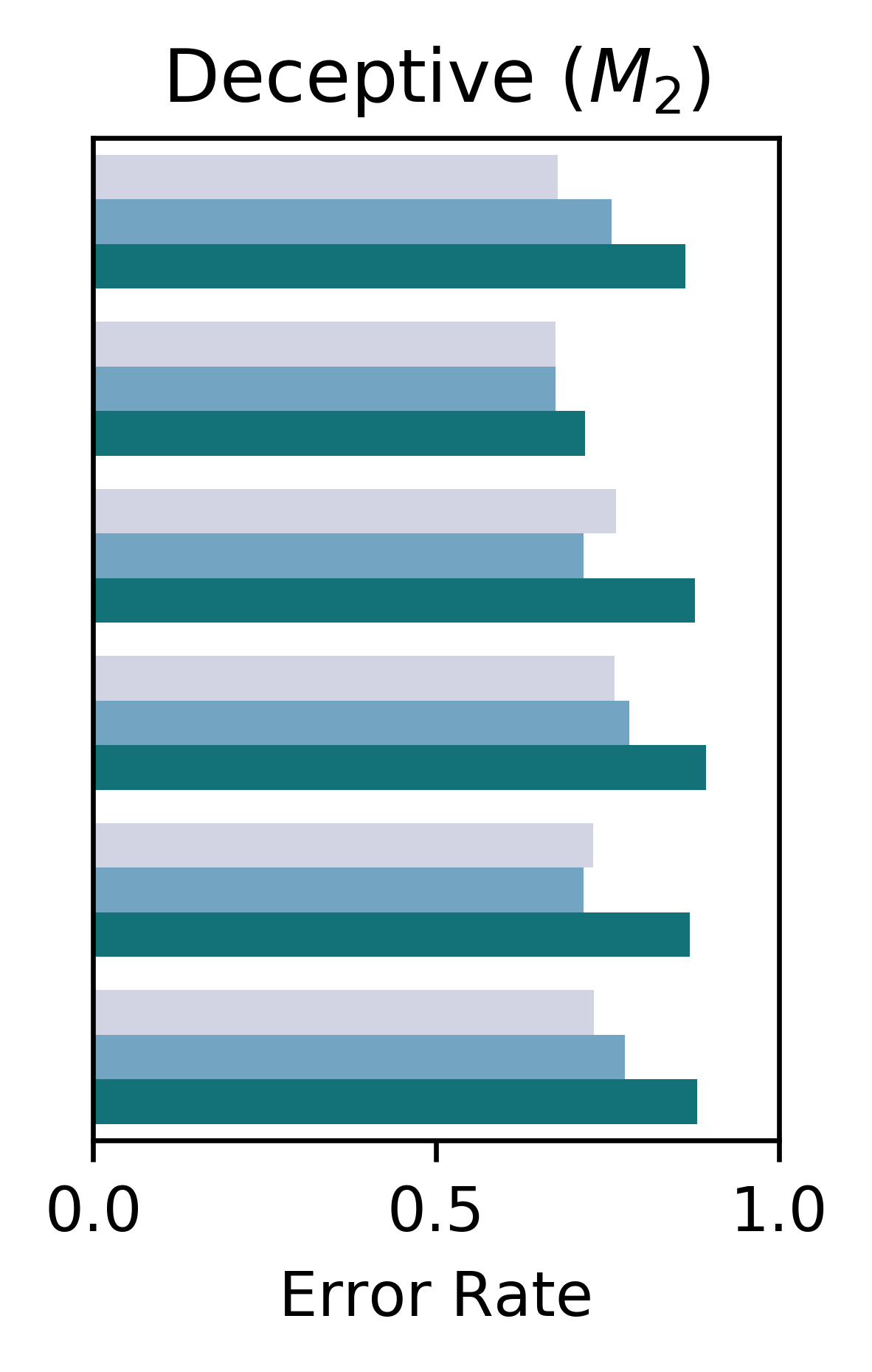}
    \includegraphics[height=8.5\baselineskip]{ 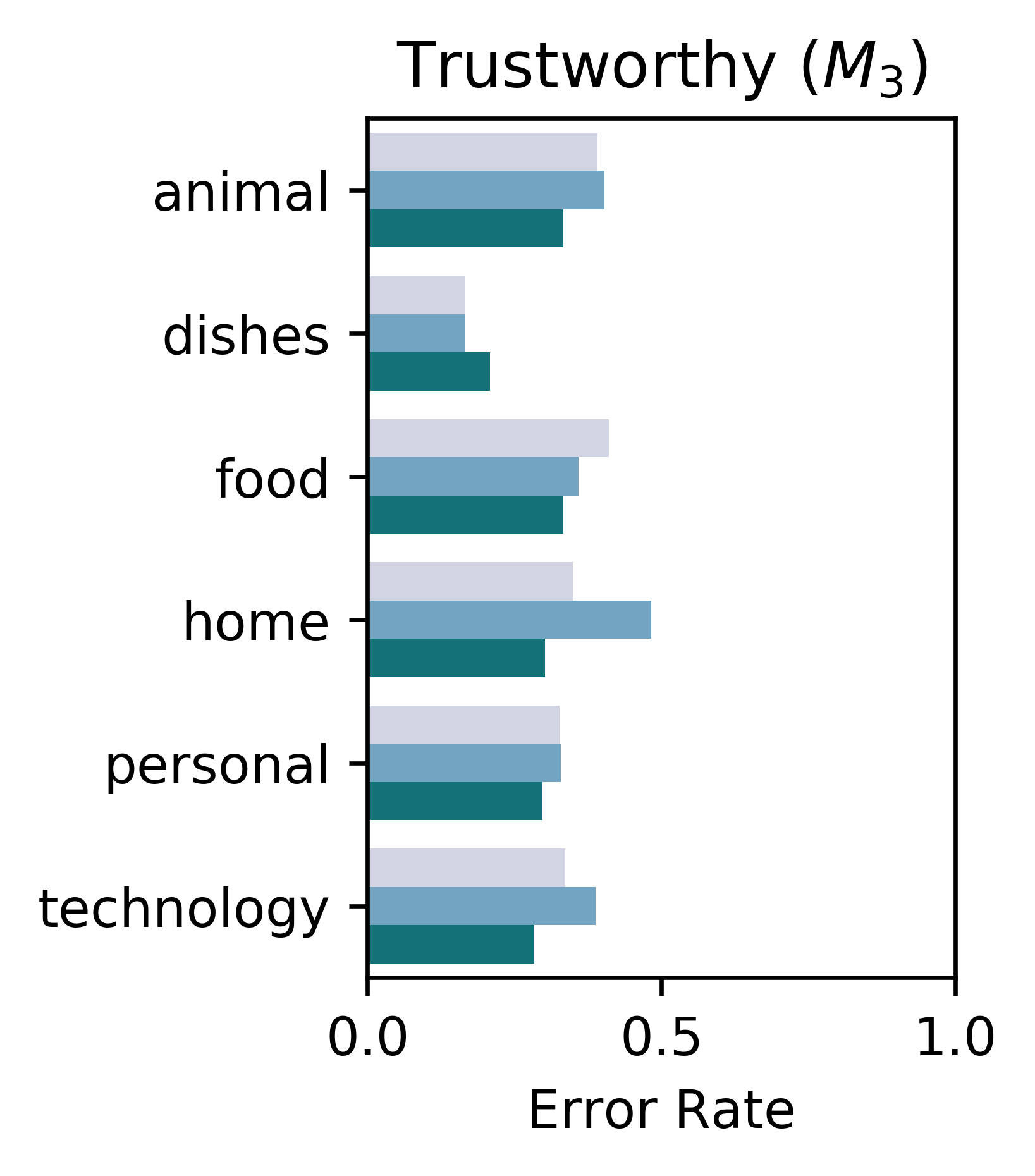}
    \includegraphics[height=8.5\baselineskip]{ 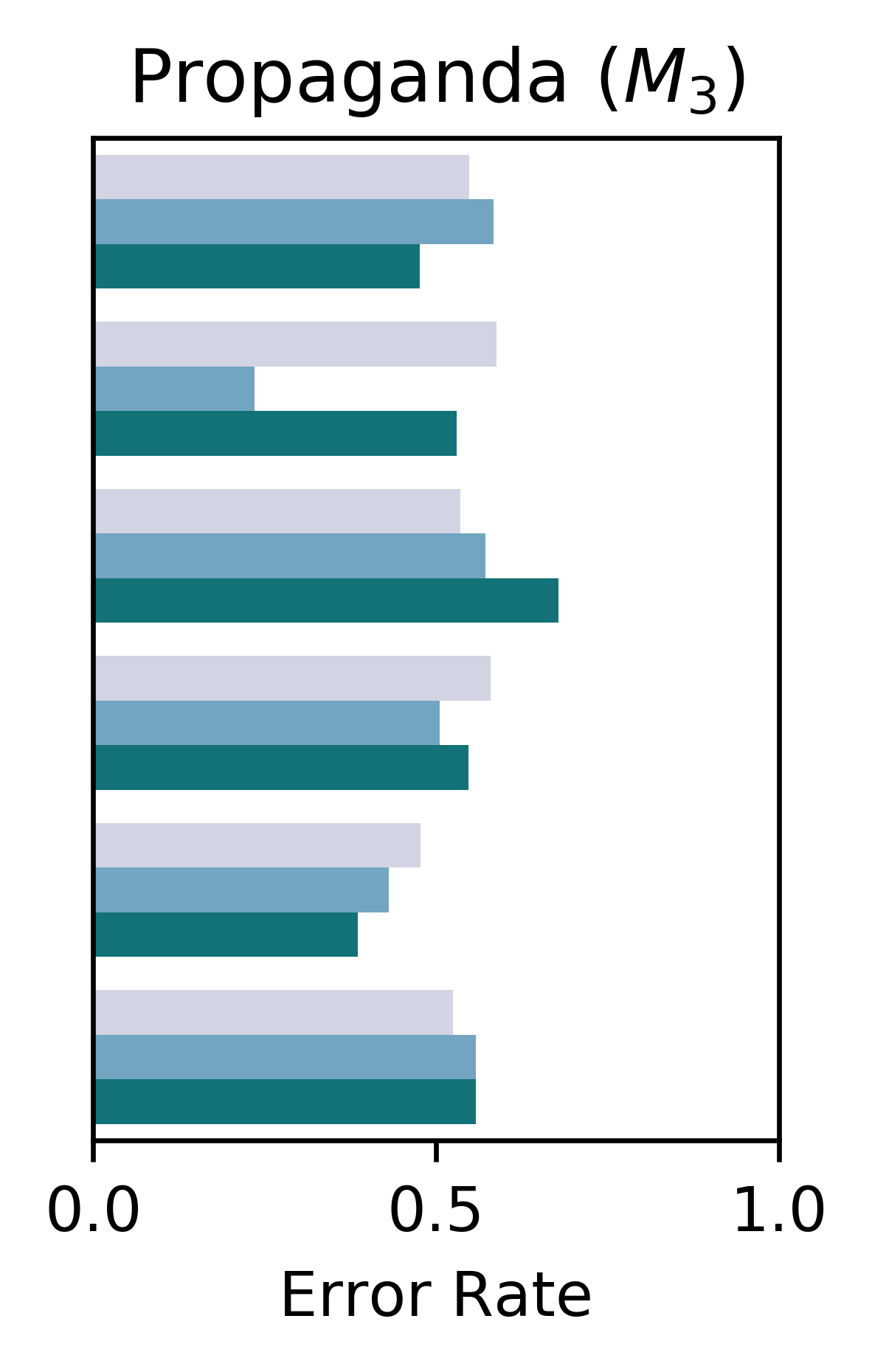}
    \includegraphics[height=8.5\baselineskip]{ 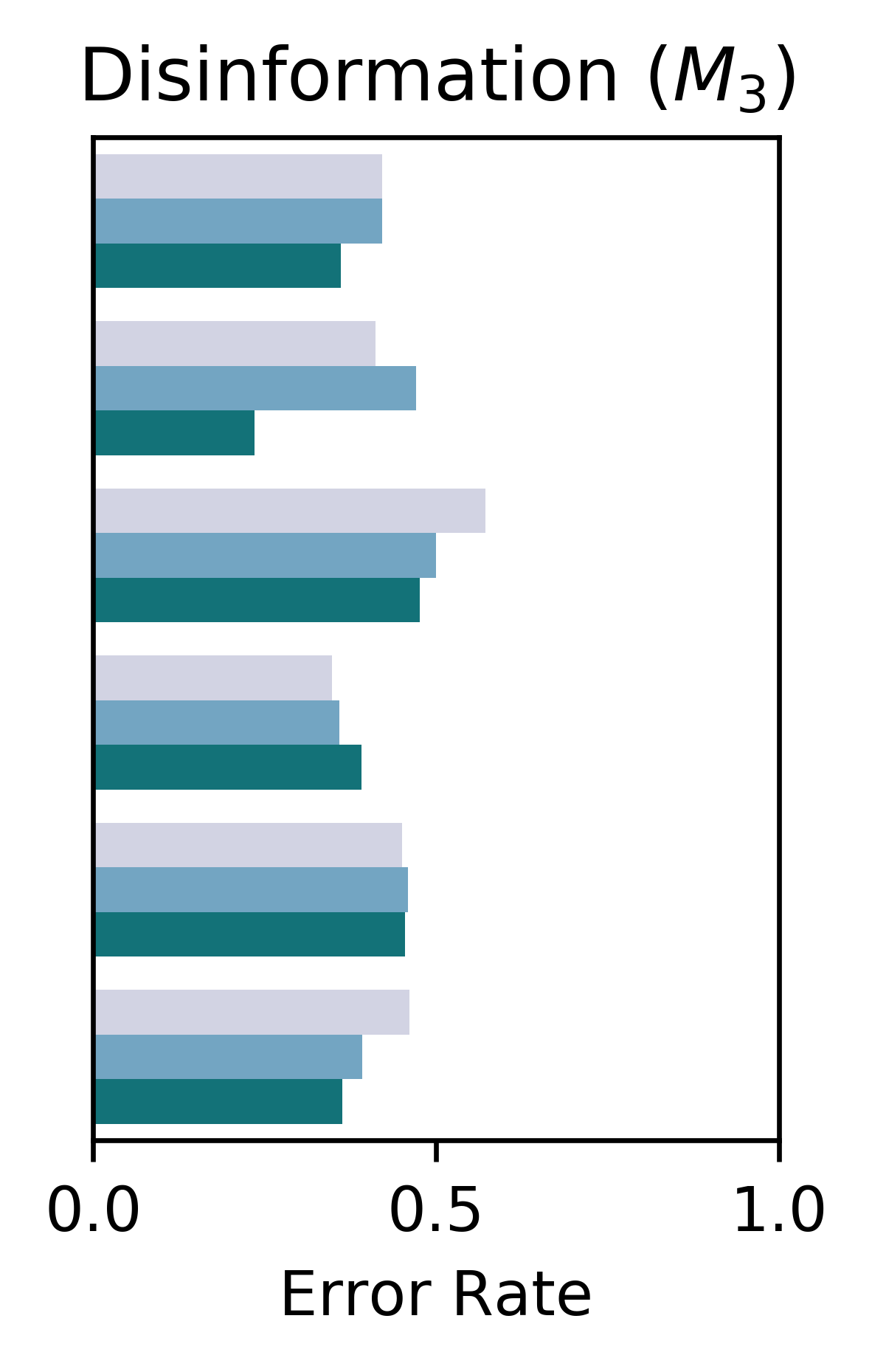}

    \includegraphics[height=8.5\baselineskip]{ 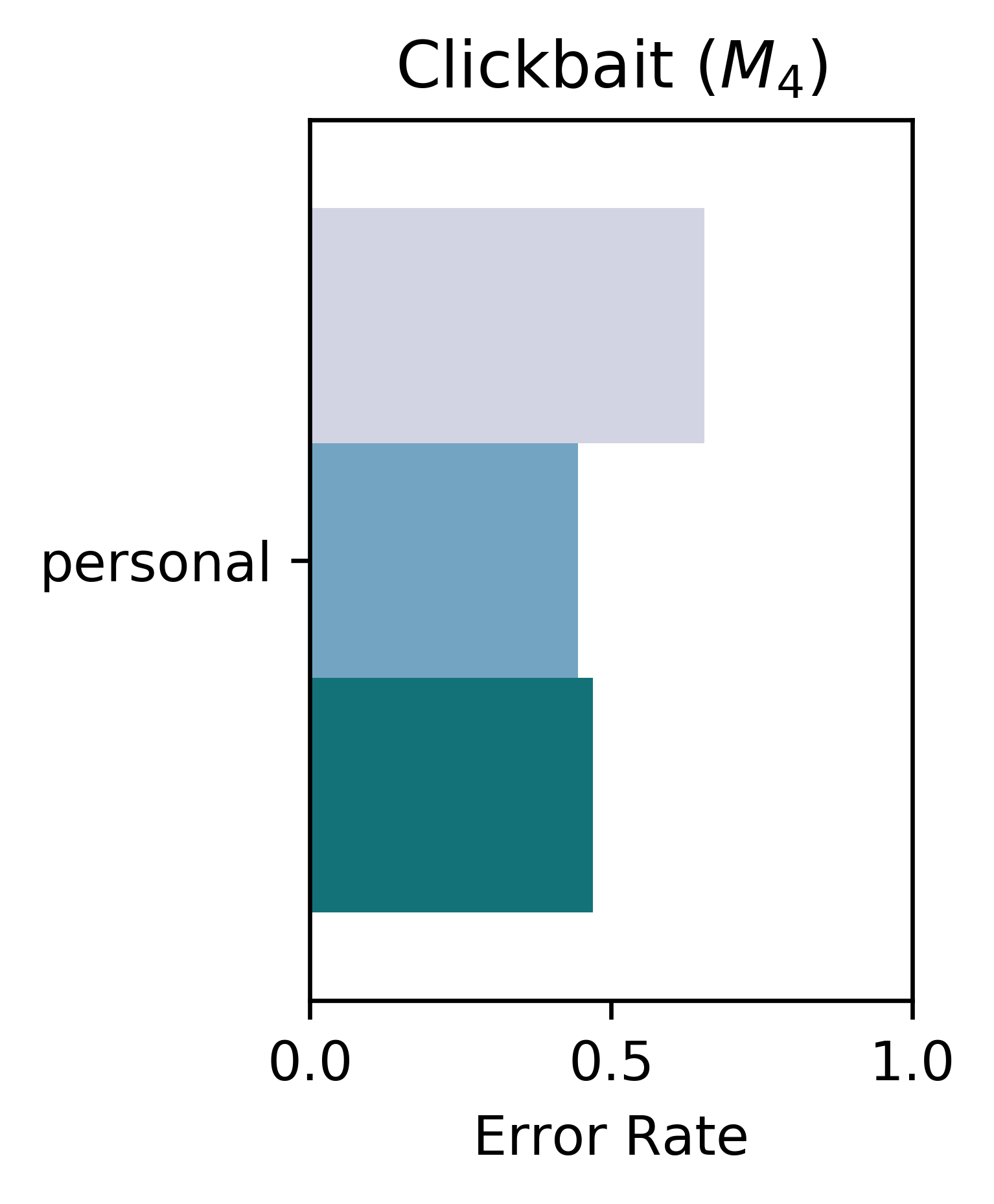}
    \includegraphics[height=8.5\baselineskip]{ 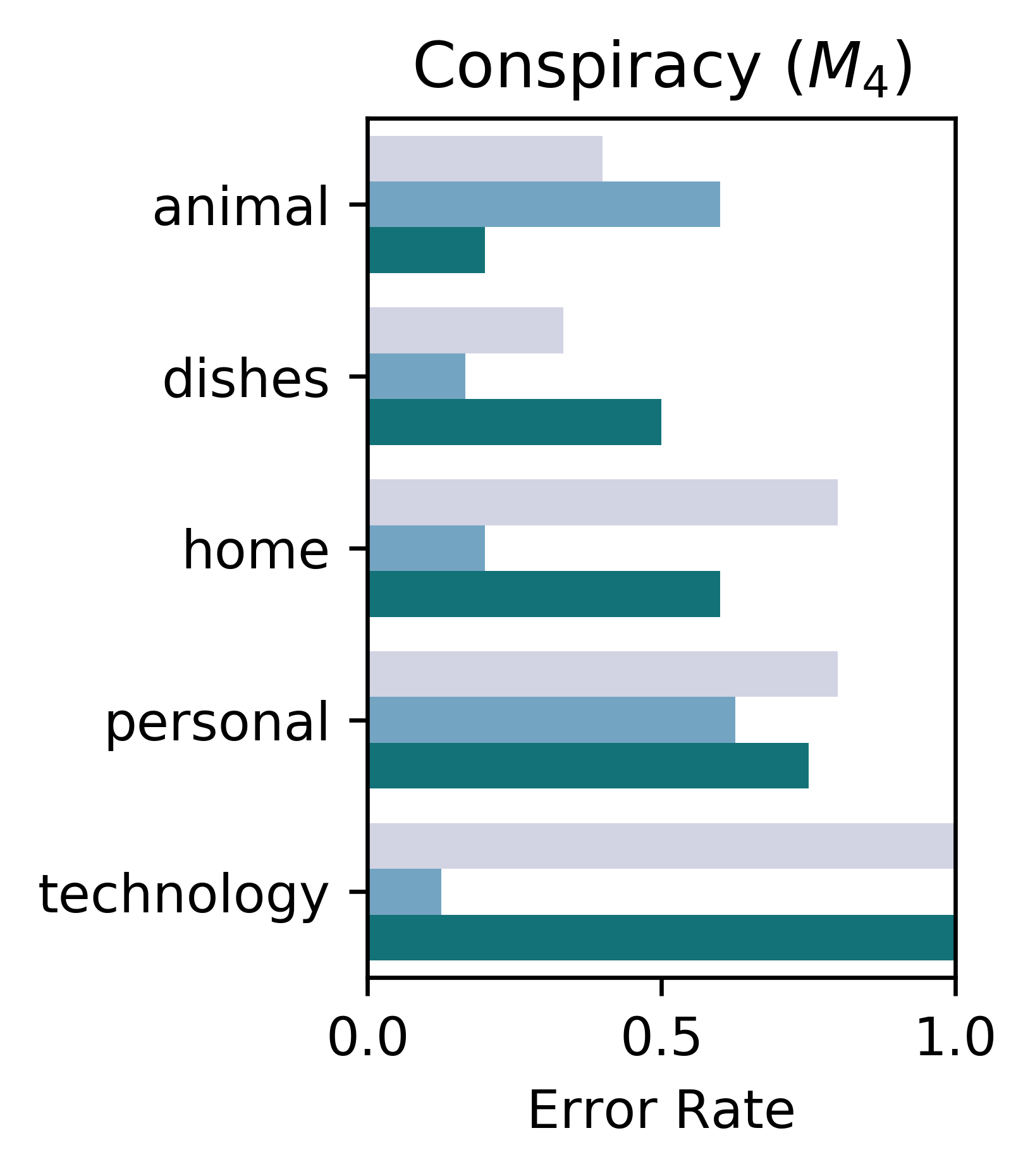}
    \includegraphics[height=8.5\baselineskip]{ 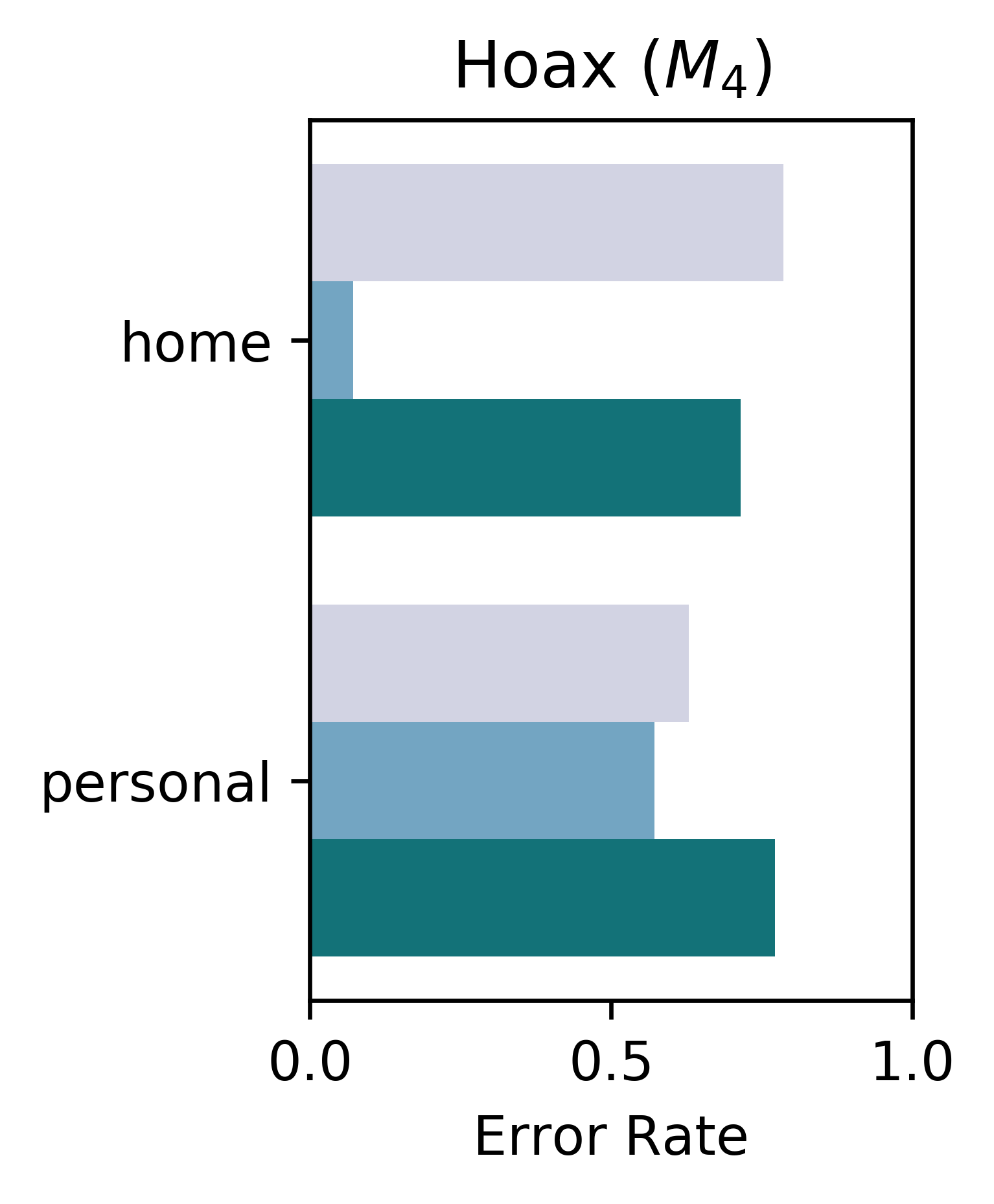}
    \includegraphics[height=8.5\baselineskip]{ 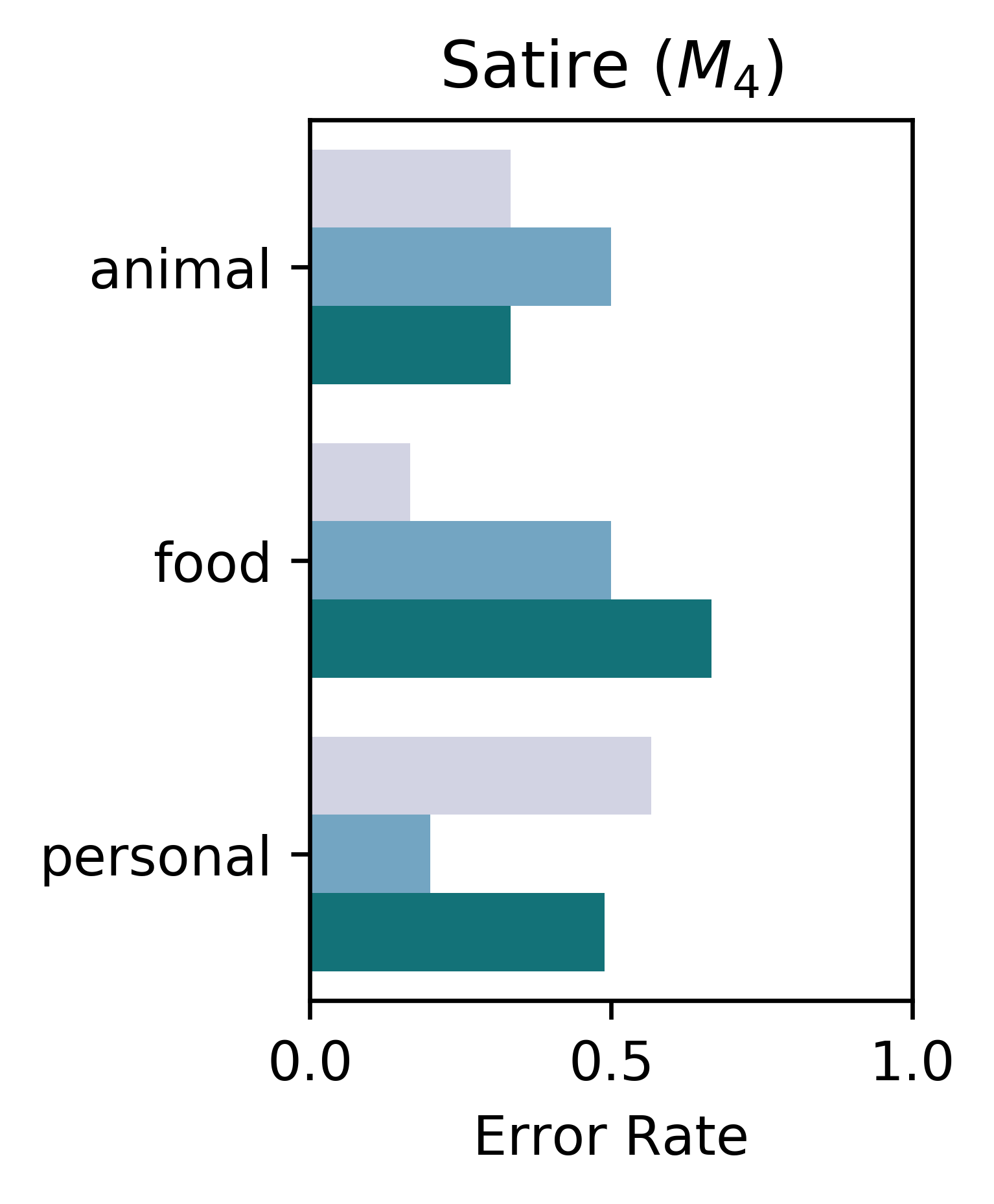}
    
    \includegraphics[height=0.5cm]{ 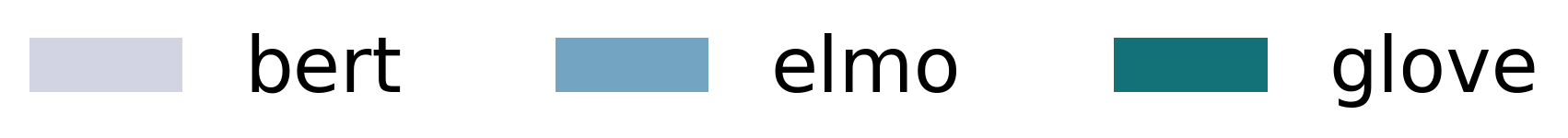}
    
    \vspace{-0.75\baselineskip}
    \caption{Error rates for Cross-Modality models for each of the six categories of primary COCO objects detected where there were at least five true class examples for which to calculate error rates.}
    \label{fig:modalities}
\end{figure}

Next, we examine the performance of our multi-modal models, presented in Figure~\ref{fig:modalities}, across each of the three tasks and two modalities: different state-of-the-art text embeddings used by the model to represent the text component of the news post and the most distinguishable objects present in the image component of the news post to be classified.  
We observe several significant findings related to these modality-specific features. 
Because of the numerous COCO~\cite{lin2014microsoft} classes, we group similar objects into super-categories\footnote{\url{https://tech.amikelive.com/node-718/what-object-categories-labels-are-in-coco-dataset/}}. For example, the {\it personal} category contains people, umbrellas, backpacks, etc.

\begin{figure}[t!] 
	\centering
	\small
	
	``death by hawthorn the bath lotion that has killed over 70 russians''
	
	\includegraphics[width=0.2\textwidth, 
	%width=0.3\textwidth, 
		 clip,keepaspectratio]{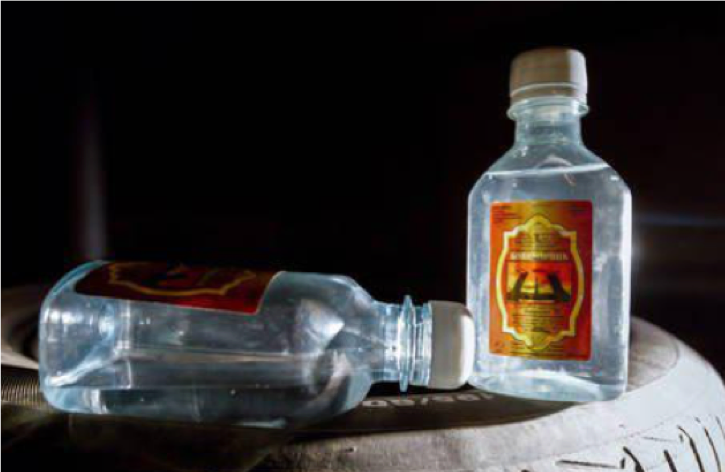}

	\caption{ An example of a disinformation tweet that the GloVe ($M_3$) model is able to correctly identify (55.17\% confidence) but the BERT and ELMo ($M_3$) models misclassify as propaganda (confidences of 48.29\% and 60.86\%).
	}
	
	\label{disinfo} 
\end{figure}

When we examine the performance of the binary cross-modality ($M_2$) model, we see the best model performance on trustworthy tweets -- with about 50\% fewer misclassifications than on deceptive tweets -- consistently across every object class and embedding type. 
In particular, the model that utilizes the pre-trained GloVe embeddings achieves the lowest error rate of 19.35\%. We also find that the model using the GloVe embeddings has the best performance --- the lowest error rate --- among the 3-way ($M_3$) models with an error rate of 23.53\%.
Figure \ref{disinfo} illustrates an example tweet where the GloVe-based ($M_3$) model has correctly classified the tweet as disinformation but the models that rely on the BERT and ELMo text embeddings to represent the text component of the tweet incorrectly classify it as a propaganda tweet.

\begin{figure}
\small
    \centering

    ``rt a reading society embraces
civilized values is adaptable'' 

	\includegraphics[width=0.15\textwidth, 
		 clip,keepaspectratio]{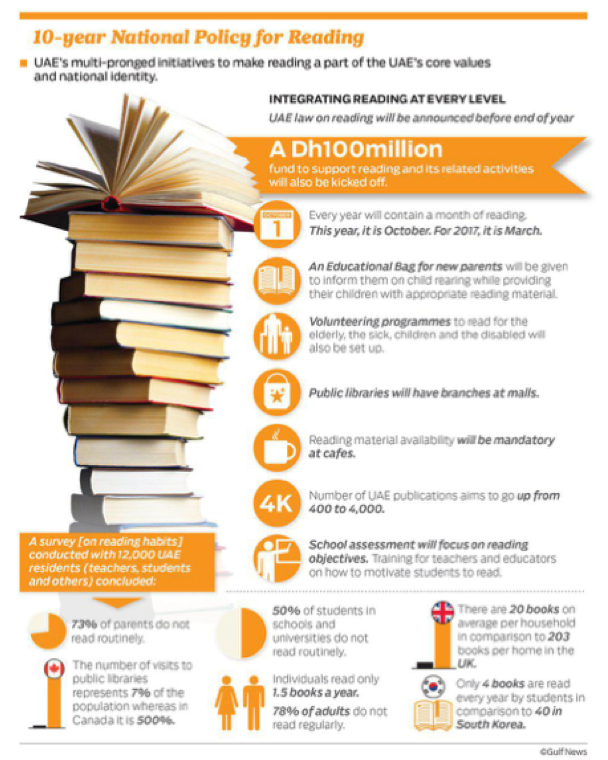}

    \caption{The text and image from a trustworthy tweet. The ELMo model misclassifies (as disinformation with 85.74\% confidence) while the BERT and GloVe models correctly classify it as trustworthy (with confidences of 76.55\% and 44.27\%).}
    \label{trust}
\end{figure}

We find that Disinformation tweets where images contain {food} items have a misclassification rate greater than 50\% for 3-way classification models ($M_3$) -- the highest of all objects. 
In contrast, trustworthy tweets have the lowest misclassification error of any class across the model types for all objects at 32.15\%. However, one anomaly is the rate of misclassification (48.19\%) of trustworthy tweets with home-related objects (such as books) from the ELMo ($M_3$) model. An example of such a tweet where this ELMo model is incorrect while the others are correct is shown in Figure \ref{trust}.
In all three classes, we see the lowest misclassification error when kitchen-related objects are present in the image. In particular, propaganda tweets, where kitchen and personal objects appear in the image, are correctly classified more often than misclassified (an error rate less than 45\%).
Interestingly, the misclassification rate is similar regardless of the embedding type (between 16.67\% and 20.08\%) for trustworthy tweets with kitchen objects. 
For propaganda tweets with kitchen objects, ELMo embeddings produce the lowest misclassification rate (23.52\%).

For the 4-way classification ($M_4$) task, we find that conspiracy tweets containing images with kitchen objects have the lowest error rate at 33.33\% while conspiracy tweets with technology have the highest error rate at 70.83\% compared to the remaining three deceptive classes.
Predictions with 
ELMo embeddings have the lowest misclassification rate (36.13\%) across all four classes and all object types compared to BERT (67.58\%) and GloVe (61.09\%). We see this particularly in cases where the BERT and GloVe models have high error rates, \eg conspiracy tweets that contain technology-related objects and hoax tweets that include home objects.

\subsection{Cross-Language Performance}

\begin{figure}[t!]
	\input{robustness_only_figs/robust_across_languages}
	\caption{Bar plots display the number of incorrect predictions for each class (Clickbait, Conspiracy, Propaganda) and overall (All) for the multilingual (All Languages) model and single-language (Per Language) models over the five languages: English, Russian, German, Spanish, and French.  Statistically significant differences in performance are indicated with * for $p<0.05$ and ** for $p<0.01$ (MWU).}%Mann Whitney U).}
	\label{fig:cross_languages_results}
	\vspace{-\baselineskip}
\end{figure}
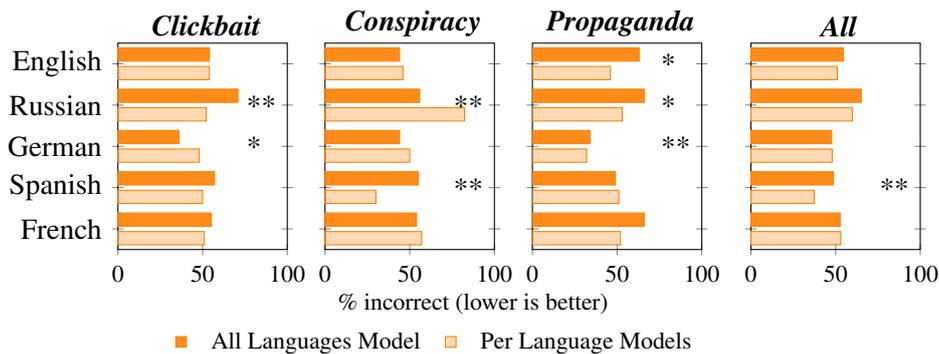

Finally, we illustrate the performance of the six cross-languages models in Figure~\ref{fig:cross_languages_results} (in these plots, lower bars indicated higher performance): a single multilingual model trained jointly on multiple languages ({\it All Languages Model}) and five models trained on each of the languages individually ({\it Per Language Models}, one for each of the five languages). As shown in Figure~\ref{fig:cross_languages_results}, the multilingual model trained jointly on multiple languages produces incorrect classifications most often on tweets that were labeled as {propaganda} when tested on {English} data while both the multilingual and single-language models tested on English data experience a steady number of errors over clickbait and conspiracy. 

When we compare the jointly trained `All Languages' model versus individually trained `Per-Language' models across the other languages, we see that the single-language (`Per Language') models trained separately with English, Russian, German, and Spanish outperform the aggregated-language model tested on the same languages with regards to propaganda. The single-language model trained and tested on Spanish data displays the best performance on conspiracy. Analytically, we can see the difference in model performance when looking at each class individually and over all classes collectively. The aggregated-language model tested with German data shows better performance over clickbait and conspiracy while making more errors with propaganda. However, the single-language model trained and tested on German data sees lesser performance over clickbait and conspiracy while making fewer errors with propaganda. When we review the aggregated- and single-language German models over all classes, they display a similar performance.

\section{Discussion and Future Work}

 We have presented an extensive evaluation of the robustness of digital deception models across frequently encountered real-world scenarios that would be necessary to benchmark against to identify reasonable performance for models considered for deployment. Our analyses have identified several trends in behavior across multiple robustness tasks and granularities of deception detection tasks (from binary to 4-way classification). To the best of our knowledge, we are the first to present this type of evaluation of deception detection model robustness. 
 
In our robustness analyses detailed above, we have illustrated the danger of relying on single performance measurement, metrics, or analyses by showcasing how a model achieving optimal performance or significant performance increases on a specific task, domain, or context (\eg multilingual, multimodal) may significantly under-perform with slight alterations in scope or context and other frequently encountered factors in real-world applications. 
 Further, we have shown several ways that out-of-domain, multilingual, and multiple modality inputs affect model performance and 
 potential methods to mitigate the impact. For example, the out-of-domain analyses showed that a domain-diverse set of training examples led to higher performance on both domains in held out test sets compared to models trained on domain-specific examples. 
 Additionally, the cross-modality analyses illustrated how frequently human-like classification mistakes can be made and which areas of the data distribution are not well represented, \eg clickbait tweets with images containing non-personal related objects. 
Results highlighted in the current work open several avenues of future work to develop, evaluate, and understand neural deception detection models in the context of real-world applications.

Future work will investigate the performance of a variety of such domain diverse training data compared to domain-specific models to identify if this hypothesis holds or if there may be other confounding factors related to specific platforms. 
In the current work, we have focused on quantifying the performance using a simplified, shared architecture that has been frequently used in deception detection models. However, future work can leverage our interactive Jupyter notebooks that will be made available at publication for reproducible extensions using our analysis framework to benchmark the performance of more complex, or newly developed state-of-the-art neural architectures.

\section*{Acknowledgements}
This research was supported by the Laboratory Directed Research and Development Program at Pacific Northwest National Laboratory, a multiprogram national laboratory operated by Battelle for the U.S. Department of Energy.

\bibliographystyle{coling}
%\bibliography{references}

\end{document}

%% file: robustness_only_figs/robustness_overview_table.tex
\small
\begin{tabular}{c @{\hspace{8pt}} l @{\hspace{8pt}} l l l |c}%@{\hspace{5pt}}c@{\hspace{5pt}}c} 
	\hline
	%&&&&& \multicolumn{3}{c}{Robustness Tasks}  \\
	
	Model & Platform(s) & Input & Text Embeddings & Detection Task (Classes) &  Robustness Task\\%\it  Domains &  \it Modalities  &  \it Languages  \\
	
	\hline 
	 \multirow{2}{*}{$M_1$}   & Twitter &  \multirow{2}{*}{Text + LC}  & \multirow{2}{*}{BERT}   
	&  \multirow{2}{*}{trustworthy, deceptive} &  \multirow{2}{*}{Cross-Domain}   \\%&   \\ 
	  & Reddit  &    &  	& \\%  &  &   \\ 
	
	%\vspace{-0.5\baselineskip}	\multicolumn{8}{c}{\cellcolor{gray}}\\
	
	\rowcolor{lightgray!20} 
	 &   & &  &    &\\% &   \\
	\rowcolor{lightgray!20} 
	\multirow{-2}{*}{$M_2$}  & \multirow{-2}{*}{Twitter} & \multirow{-2}{*}{Text + Image + LC}  & \multirow{-2}{*}{BERT, GLoVe, ELMo}  & \multirow{-2}{*}{trustworthy, deceptive}  &  \multirow{-2}{*}{Cross-Modality}\\%& \multirow{-2}{*}{X}       \\ 

	%%\multirow{2}{*}{$M_3$} & \multirow{2}{*}{Twitter} & \multirow{2}{*}{Text + Image} & \multirow{2}{*}{BERT, GLoVe, ELMo}  & \multirow{2}{*}{trustworthy, propaganda, disinformation} & & \multirow{2}{*}{X} &\multirow{2}{*}{X} \\
	\multirow{2}{*}{$M_3$} & \multirow{2}{*}{Twitter} & \multirow{2}{*}{Text + Image + LC} & \multirow{2}{*}{BERT, GLoVe, ELMo}  & \multirow{1}{*}{trustworthy, propaganda,} & \multirow{2}{*}{Cross-Modality}\\%& \multirow{2}{*}{X}   \\
	  &   & &  &   disinformation &\\% &  \\
	  %&   & &  &    & &    \\

	  \rowcolor{lightgray!20} 
	    &   &   &   & \multirow{1}{*}{clickbait, satire,} &\\% &   \\ 
	  \rowcolor{lightgray!20} 
	   \multirow{-2}{*}{$M_4$} & \multirow{-2}{*}{Twitter} & \multirow{-2}{*}{Text + Image + LC} & \multirow{-2}{*}{BERT, GLoVe, ELMo} & \multirow{1}{*}{hoax, conspiracy} & \multirow{-2}{*}{Cross-Modality}\\%&   \multirow{-2}{*}{X}  \\

	    &   &   &   & \multirow{1}{*}{clickbait, conspiracy,} &\\% &   \\ 
	   \multirow{-2}{*}{$M_5$} & \multirow{-2}{*}{Twitter} & \multirow{-2}{*}{Text} & \multirow{-2}{*}{Character} & \multirow{1}{*}{propaganda} & \multirow{-2}{*}{Cross-Language}\\%&   &\multirow{-2}{*}{X}  \\
	  
		%&  3-way: clickbait, conspiracy, propaganda & LSTM & Text (Russian, English, German, Spanish, French)  & Character
		
	\hline

	\hline
	
\end{tabular}

\ignore{
\begin{tabular}{c l l l l |c c c} 
	\hline
	&&&&& \multicolumn{2}{c}{Robustness} & Adversarial \\
	
	Model & Platform(s) & Input & Text Embeddings & Task (Classes) &  \it Domains &  \it Modalities & ~ \\
	\hline 
	%% M1:
	 \multirow{2}{*}{$M_1$}   & Twitter &  \multirow{2}{*}{Text}  & \multirow{2}{*}{BERT}   
	&  \multirow{2}{*}{trustworthy, deceptive} &  \multirow{2}{*}{X}   & &  \\ 
	  & Reddit  &    &  	&   &  &  & \\ 
	 
	%% M2: 
	\rowcolor{lightgray!20}
	  &   & &  &    ~ & &  & \\
	\rowcolor{lightgray!20}
	\multirow{-2}{*}{$M_2$}  & \multirow{-2}{*}{Twitter} & \multirow{-2}{*}{Text + Image}  & \multirow{-2}{*}{BERT, GLoVe, ELMo}  & \multirow{-2}{*}{trustworthy, deceptive}  &  & \multirow{-2}{*}{X}   &      \\
	
	 %M3:
	\multirow{2}{*}{$M_3$} & \multirow{2}{*}{Twitter} & \multirow{2}{*}{Text + Image} & \multirow{2}{*}{BERT, GLoVe, ELMo}  & \multirow{2}{*}{trustworthy, propaganda, disinformation} & & \multirow{2}{*}{X} &\multirow{2}{*}{X} \\
	  &   & &  &   ~ & &  & \\ 
	  
	 %M4:
	\rowcolor{lightgray!20}
	  &   &   &   &    ~ & &  & \\ 
	  \rowcolor{lightgray!20} 
	  \multirow{-2}{*}{$M_4$} & \multirow{-2}{*}{Twitter} & \multirow{-2}{*}{Text + Image} & \multirow{-2}{*}{BERT, GLoVe, ELMo} & \multirow{-2}{*}{clickbait, satire, hoax, conspiracy} & &  \multirow{-2}{*}{X} & \multirow{-2}{*}{X} \\
	\hline

\end{tabular}
}

\ignore{ %%% Old table
\begin{tabular}{@{\hskip .2cm} l@{\hskip .2cm} l @{\hskip .2cm} l @{\hskip .2cm} l @{\hskip .2cm} ll p{3cm} }%p{6cm} l p{3.3cm} l}
	%{l p{6cm} l p{3.3cm} l}
	%{llllll}
	%{l p{3.7cm} p{3.5cm} p p{3.5cm} l}
		\hline
		\multicolumn{2}{l}{Analysis} 
		%Analysis 
		& Platform(s) & Model & Input & Text Embeddings & Classes\\
		\hline
		
		\multirow{6}{*}{Robustness} &
		\multirow{2}{*}{Cross-domains}  
		 &  Twitter %Twitter~\cite{volkova2019explaining}~
		 & \multirow{2}{*}{M1}  & \multirow{2}{*}{Text (English)}  & \multirow{2}{*}{BERT}  & trustworthy, deceptive\\
		 &&   
		 Reddit %Reddit~\cite{glenski2018identifying} 
		 &    &    &    \\
		\cline{2-6}%\hline
		
		\ignore{
		%Robustness 
		&
		Cross-languages
		&   Twitter
		%\multirow{2}{*}{Cross-languages} 
		%&   \multirow{2}{*}{Twitter}   
		%%\multirow{2}{*}{Twitter~\cite{glenski2019multilingual}}  
		& M3  & Text (Russian, English, German, Spanish, French)  & Character \\
		\cline{2-6}%\hline
	}
		%\multirow{3}{*}{Robustness} 
		&
		\multirow{3}{*}{Cross-modalities} 
		& 
		\multirow{3}{*}{Twitter}  
		%\multirow{3}{*}{Twitter~\cite{volkova2019explaining}}  
		&  M2  & Text (English) + Image & BERT, GLoVe, ELMo  & trustworthy, deceptive \\
		&&   & M4  & Text (English)+ Image  & BERT, GLoVe, ELMo & trustw., propag., disinfo \\
		&&  & M5   & Text (English) + Image & BERT, GLoVe, ELMo \\
		\hline

		\multicolumn{2}{l}{\multirow{2}{*}{Adversarial} }
		& 
		\multirow{2}{*}{Twitter}  
		& M4  & Text (English)+ Image  & BERT, GLoVe, ELMo & trustw., propag., disinfo\\ 
		&&
		& M5   & Text (English) + Image & BERT, GLoVe, ELMo & clickb., satire, hoax,conspir.\\  
		%&  & M2  & Text (English) + Image & BERT, GLoVe, ELMo \\
		\hline        
		
	\end{tabular}
}

\ignore{
	\begin{tabular}{l p{3.7cm} p{3.5cm} p{3.5cm} l}
		\hline
		Analysis & Task & Model Architecture & Input & Text Embeddings\\
		\hline
		
		\multirow{2}{*}{Cross-domains}  
		& 2-way: trustworthy, deceptive & LSTM + Lex & Text (English)  & BERT  \\
		\hline
		\multirow{2}{*}{Cross-languages} 
		&  3-way: clickbait, conspiracy, propaganda & LSTM & Text (Russian, English, German, Spanish, French)  & Character \\
		\hline
		\multirow{6}{*}{Cross-modalities} 
		& 4-way: clickbait, hoax, satire, conspiracy  & LSTM + ResNet Vec + Lex  & Text (English) + Image & BERT, GLoVe, ELMo \\
		& 3-way: trustworthy, disinformation, propaganda & LSTM + ResNet Vec + Lex & Text (English)+ Image  & BERT, GLoVe, ELMo \\
		& 2-way: trustworthy, deceptive & LSTM + ResNet Vec + Lex & Text (English) + Image & BERT, GLoVe, ELMo \\
		\hline        
\end{tabular}
}

%% file: figs/exp1confusionmatrix.tex
\centering
\small

 \begin{tabular}{@{\hspace{4pt}}c@{\hspace{10pt}}c@{\hspace{4pt}}r@{\hspace{10pt}}rr@{\hspace{7pt}}c@{\hspace{7pt}}rr}  
 %	\begin{tabular}{@{\hspace{1pt}}c@{\hspace{3pt}}c@{\hspace{1pt}}rrrcrr}  
		 %& 
		 %&&& \multicolumn{2}{c}{ \bf Predicted Class}
		 %&&\multicolumn{2}{c}{ \bf Predicted Class} \\
		 &&&\multicolumn{5}{c}{ \bf Predicted Class} \\
		 
		  & & &  Trustworthy 
		  &   Deceptive~ 
		  & & 
		   Trustworthy 
		  &  Deceptive~
		   \\

		 \multirow{3}{*}{\it  Train on}  
		 
		 & %\parbox[t]{2mm}{\multirow{4}{*}{\rotatebox[origin=c]{90}{\tiny \bf True Class}}}
		 & \multirow{2}{*}{  Trustworthy } %{ \tiny Trustworthy} 
		 & 46\% \cellcolor{correct!46} & 54\%  \cellcolor{incorect!54}& 
		 & 89\% \cellcolor{correct!89}& 11\% \cellcolor{incorect!11} 
		 \\

		  & 
		  &  %\tiny (Trustw.)
		  & \tiny  (1.72k)  \cellcolor{correct!46} & \tiny (2.02k)  \cellcolor{incorect!54}& 
		  &\tiny  (4.77k)  \cellcolor{correct!89}& \tiny (573)  \cellcolor{incorect!11} 
		  \\

		    \textit{\textbf{Twitter}} &
		    & \multirow{2}{*}{Deceptive}% \tiny Deceptive % \multirow{2}{*}{ \tiny Deceptive}  %\tiny Deceptive
		    &  19\%   \cellcolor{incorect!19} & 
		    81\% \cellcolor{correct!81}& 
		    & 90\% \cellcolor{incorect!90}& 
		    10\%  \cellcolor{correct!10}
		    \\ 
		  
		  &
		  & %  \tiny (Deceptivecep.)
		  & \tiny  (1.38k)  \cellcolor{incorect!19} 
		  &  \tiny  (5.79k) \cellcolor{correct!81}& 
		  &  \tiny  (4.72k) \cellcolor{incorect!90}& 
		  \tiny  (517)  \cellcolor{correct!10}
		  \\

		  \rule{0pt}{0ex}\\[-1.2ex]%.5ex] %[-2ex]

		  \multirow{3}{*}{ \it  Train on}  
		  
		  & \parbox[t]{2mm}{\multirow{4}{*}{\rotatebox[origin=c]{90}{ \bf True Class}}}
		  & \multirow{2}{*}{ Trustworthy} %\tiny Trustworthy} 
		  & 87\%  \cellcolor{correct!87}
		  & 13\%  \cellcolor{incorect!13}& 
		  & 62\% \cellcolor{correct!62}
		  & 38\%  \cellcolor{incorect!38} 
		  \\ 
		    
		  & 
		  &  
		  & \tiny  (3.27k)  \cellcolor{correct!87}
		  & \tiny  (469) \cellcolor{incorect!13}& 
		  & \tiny  (3.28k) \cellcolor{correct!62}
		  & \tiny  (2.06k)  \cellcolor{incorect!38} 
		  \\

		  \textit{\textbf{Reddit} }&
		  & \multirow{2}{*}{ Deceptive }%\tiny Deceptivecep.}   
		  & 89\% \cellcolor{incorect!89}& 11\% \cellcolor{correct!11}& 
		  & 51\% \cellcolor{incorect!51}& 49\% \cellcolor{correct!49}
		  \\  
		  
		  &
		  &  
		  &\tiny  (6.41k)  \cellcolor{incorect!89}
		  & \tiny  (757) \cellcolor{correct!11}& 
		  & \tiny  (2.69k) \cellcolor{incorect!51}
		  & \tiny  (2.55k)  \cellcolor{correct!49}
		  \\

		  % \ignore{
		   \rule{0pt}{0ex}\\[-1.2ex]%.5ex] %[-2ex]
		   
		   \multirow{3}{*}{ \it Train on}  
		   
		   & %\parbox[t]{2mm}{\multirow{4}{*}{\rotatebox[origin=c]{90}{\tiny \bf True Class}}}
		   & \multirow{2}{*}{ Trustworthy } %\tiny Trustworthy} 
		   & 
		   60\%	﻿\cellcolor{correct!60	}&	40\%	﻿ \cellcolor{incorect!40	}&	&	59\%	﻿\cellcolor{correct!59	}&	41\%	﻿ \cellcolor{incorect!41	}\\

		   & 
		   &  
		   &  
		   \tiny  (2.26k)		﻿\cellcolor{correct!60	}
		   &		 \tiny  (1.49k)		﻿ \cellcolor{incorect!40	}
		   &	
		   &		 \tiny  (3.22k)		﻿\cellcolor{correct!59	}
		   &		 \tiny  (2.21k)		﻿ \cellcolor{incorect!41	}\\

		   \textit{\textbf{Twitter}} &
		   & \multirow{2}{*}{ Deceptive } % \tiny Deceptive}   
		   &	
		   20\%	﻿ \cellcolor{incorect!20	}&	80\%	﻿\cellcolor{correct!80	}&	&	28\% \cellcolor{incorect!28	}&	72\%	﻿\cellcolor{correct!72	}\\
		   
		   \bf \textit{\textbf{+ Reddit}}&
		   &  
		   & 
		   \tiny  (1.43k)		﻿ \cellcolor{incorect!20	}&		\tiny  (5.75k)		﻿\cellcolor{correct!80	}&	&		\tiny  (1.55k)		﻿ \cellcolor{incorect!28	}&		\tiny  (3.95k)		﻿\cellcolor{correct!72	}\\
		%}
		  
		  \rule{0pt}{0ex}\\[-1ex]%.25ex]

		    %&  
		    %&  &   & \multicolumn{2}{c}{Test on \textbf{Twitter}} & & \multicolumn{2}{c}{Test on \textbf{Reddit}}   \\ 
		    &  &   & \multicolumn{2}{c}{\it Test on} & & \multicolumn{2}{c}{\it Test on}   \\ 
		    &  &   & \multicolumn{2}{c}{\textit{\textbf{Twitter}}} & & \multicolumn{2}{c}{\textit{\textbf{ Reddit}}}   \\ 
		    
	\end{tabular} 

%% file: robustness_only_figs/robust_across_languages.tex
\centering 

\begin{tikzpicture}
\begin{axis}[ 
xbar,
title =\textit{ \textbf{Clickbait}}, 
height = 1.7in,  width = 1.5in, %1.2in, 
xtick align = center, ytick align = center,
ylabel style = {yshift = -.15in}, 
title style = {yshift = -.1in,align = center}, 
xtick pos = left,   
ymin=-0.5,ymax=4.5, 
ytick={0,1,2,3,4},
yticklabels={French, Spanish, German, Russian, English},
yticklabels={fr, es, de, ru, en},
yticklabels={French, Spanish, German, Russian, English},
xmin=0,xmax=1,
xtick={0,0.5,1}, xticklabels={0, 50, 100},
xticklabel style={font=\small, %rotate=30,
anchor=north},% east},
bar width =5,
]

 \addplot[crosslangPerLangModel, crosslangPerLangModelbar] table [col sep=comma,x=clickbaitPerLangModel, y=langorder]{figs/robust_multilingual_models.csv};
 
 \addplot[crosslangAllLangModel, crosslangAllLangModelbar] table [col sep=comma,x=clickbaitAllLangModel, y=langorder]{figs/robust_multilingual_models.csv};

% fr = 0
% es = 1
% de = 2
% ru = 3
% en = 4
\node[right,align=center] at (axis cs:0.7,3) {**}; %p=0.0038
\node[right,align=center] at (axis cs:0.7,2) {*}; %p=0.043

\end{axis}
\end{tikzpicture} 
\hspace{-0.4cm}
\begin{tikzpicture}
\begin{axis}[ 
xbar,
title =\textit{ \textbf{Conspiracy}}, 
height = 1.7in,  width = 1.5in, %1.2in, 
xtick align = center, ytick align = center,
ylabel style = {yshift = -.15in}, 
title style = {yshift = -.1in,align = center}, 
xtick pos = left,   
ymin=-0.5,ymax=4.5, 
ytick={0,1,2,3,4},
yticklabels={ ,  ,  ,   ,  },
xmin=0,xmax=1,
xtick={0,0.5,1}, xticklabels={0, 50, 100},
xticklabel style={font=\small, %rotate=30,
anchor=north},% east}, 
bar width =5,
]

\addplot[crosslangPerLangModel, crosslangPerLangModelbar] table [col sep=comma,x=conspiracyPerLangModel, y=langorder]{figs/robust_multilingual_models.csv};

\addplot[crosslangAllLangModel, crosslangAllLangModelbar] table [col sep=comma,x=conspiracyAllLangModel, y=langorder]{figs/robust_multilingual_models.csv};

% fr = 0
% es = 1
% de = 2
% ru = 3
% en = 4
\node[right,align=center] at (axis cs:0.7,3) {**}; %p=0.00044
\node[right,align=center] at (axis cs:0.7,1) {**}; %p=0.00018

\end{axis}
\end{tikzpicture} 
\hspace{-0.3cm}%25cm}
\begin{tikzpicture}
\begin{axis}[ 
xbar,
title =\textit{ \textbf{Propaganda}}, 
height = 1.7in,  width = 1.5in, %1.2in, 
xtick align = center, ytick align = center,
ylabel style = {yshift = -.15in}, 
title style = {yshift = -.1in,align = center}, 
xtick pos = left,   
ymin=-0.5,ymax=4.5, 
ytick={0,1,2,3,4},
yticklabels={ ,  ,  ,   ,  },
xmin=0,xmax=1,
xtick={0,0.5,1}, xticklabels={0, 50, 100},
xticklabel style={font=\small, %rotate=30,
anchor=north},% east},
bar width =5,
]

\addplot[crosslangPerLangModel, crosslangPerLangModelbar] table [col sep=comma,x=propagandaPerLangModel, y=langorder]{figs/robust_multilingual_models.csv};

\addplot[crosslangAllLangModel, crosslangAllLangModelbar] table [col sep=comma,x=propagandaAllLangModel, y=langorder]{figs/robust_multilingual_models.csv};

% fr = 0
% es = 1
% de = 2
% ru = 3
% en = 4
\node[right,align=center] at (axis cs:0.7,4) {*}; %p=0.031
\node[right,align=center] at (axis cs:0.7,3) {*}; %p=0.022
\node[right,align=center] at (axis cs:0.7,2) {**}; %p=0.008

\end{axis}
\end{tikzpicture} 
\hspace{-0.25cm}
\begin{tikzpicture}
\begin{axis}[ 
xbar,
title =\textit{ \textbf{All}}, 
height = 1.7in,  width = 1.5in, %1.2in, 
xtick align = center, ytick align = center,
ylabel style = {yshift = -.15in}, 
title style = {yshift = -.1in,align = center}, 
xtick pos = left,   
ymin=-0.5,ymax=4.5, 
ytick={0,1,2,3,4},
yticklabels={ ,  ,  ,   ,  },
xmin=0,xmax=1,
xtick={0,0.5,1}, xticklabels={0, 50, 100},
xticklabel style={font=\small, %rotate=30,
anchor=north},% east},
bar width =5,
]

% fr = 0
% es = 1
% de = 2
% ru = 3
% en = 4 

\addplot[crosslangPerLangModel, crosslangPerLangModelbar] table [col sep=comma,x=allPerLangModel, y=langorder]{figs/robust_multilingual_models.csv};

\addplot[crosslangAllLangModel, crosslangAllLangModelbar] table [col sep=comma,x=allAllLangModel, y=langorder]{figs/robust_multilingual_models.csv};
 
\node[right,align=center] at (axis cs:0.7,1) {**}; %p=0.002

\end{axis}
\end{tikzpicture} 

\vspace{-0.25\baselineskip}
\centering
\small 
\% incorrect (lower is better)

\vspace{-0.5\baselineskip}
%\centering
\begin{tikzpicture}
\begin{axis}[
%xbar,
hide axis,
height = 1in,width=1in,
xmin = 0, xmax = 50, ymin = 0, ymax = 0.4,
legend cell align = {left}, legend columns=2, legend style = {column sep=0.1in,draw=none, at = {(0,0.75)}},]
\small

\addlegendimage{only marks,  mark=square*, crosslangAllLangModel, crosslangAllLangModelbar}
\addlegendentry{All Languages Model};%Multilingual Model (All Languages)}; 
\addlegendimage{only marks, mark=square*, crosslangPerLangModel, crosslangPerLangModelbar}
\addlegendentry{Per Language Models}; 

\end{axis}
\end{tikzpicture} 

\vspace{-\baselineskip}